\documentstyle[12pt]{article}
\begin{document}

\title{Extraction of V-N-Collocations from Text Corpora: 
A Feasibility Study for German\footnote{
Revised version of the paper presented at the First ACL--Workshop on
Very Large Corpora, Columbus, Ohio, June 1993.  My thanks go to the
IdS that made the two corpora available for research purposes, to
Angelika Storrer for her steady encouragement and many fruitful
discussions, and to Mats Rooth and Matthias Heyn who introduced me to
the corpora tools.} }

\author{Elisabeth Breidt \\[1ex] 
Seminar f\"{u}r Sprachwissenschaft \\ 
Universit\"{a}t T\"{u}bingen \\ 
Wilhelmstr.\ 113, D-72074 T\"{u}bingen, Germany \\
breidt@sfs.nphil.uni-tuebingen.de \\ }

\date{October 30, 1995}

\maketitle

\begin{center}
cmp-lg/9603006
\end{center}

\begin{abstract}
The usefulness of a statistical approach suggested by Church et al.\
(1991) is evaluated for the extraction of verb--noun (V--N) collocations
from German text corpora.  Some problematic issues of that method
arising from properties of the German language are discussed and
various modifications of the method are considered that might improve
extraction results for German. The precision and recall of all variant
methods is evaluated for V--N collocations containing support verbs,
and the consequences for further work on the extraction of
collocations from German corpora are discussed.

With a sufficiently large corpus ($\ge$ 6 mio.\ word--tokens), the
average error rate of wrong extractions can be reduced to 2.2\%
(97.8\% precision) with the most restrictive method, however with a
loss in data of almost 50\% compared to a less restrictive method with
still 87.6\% precision. Depending on the goal to be achieved, emphasis
can be put on a high recall for lexicographic purposes or on high
precision for automatic lexical acquisition, in each case
unfortunately leading to a decrease of the corresponding other
variable.  Low recall can still be acceptable if very large corpora
(i.e.\ 50 -- 100 million words) are available or if corpora for special
domains are used in addition to the data found in machine readable
(collocation) dictionaries.
\end{abstract}

\section{Introduction}
Collocations present an area that is important both for lexicography
to improve their coverage in modern dictionaries as well as for
lexical acquisition in computational linguistics, where the goal is to
build either large reusable lexical databases (LDBs) or specific
lexica for specialized NLP--applications. We have tested a statistical
approach using Mutual Information (MI) and t--score, introduced into
linguistics by Church and Hanks (1989) and Church et al.\ (1991), for
the (semi--)automatic extraction of verb--noun (V--N) collocations from
untagged German text corpora. At the time when this study was carried
out in early 1993, no POS--tagged German corpora were available. In the
meantime this has changed, even work on shallow parsers for German is
in progress (Abney, University of T\"ubingen). Nevertheless it is
interesting to answer the question how much can be done with an
untagged corpus and what might be gained by lemmatizing, POS--tagging
or shallow parsing.

\section{What Do We Mean by `Collocation'?}

Collocations in the sense of `frequently cooccurring words' can quite
easily be extracted from corpora by statistic means. From a linguistic
point of view, however, a more restricted use of the term is
preferable which takes into account the difference between what
Sinclair (1966) called {\sl casual} vs.\ {\sl significant}
collocations. Casual word combinations show a normal, free syntagmatic
behaviour. In this paper, collocations shall refer only to word
combinations with a lexically (rather than syntactically or
semantically) restricted combinatory potential, where at least one
component has a special meaning that it cannot have in a free
syntagmatic construction. For example, in {\sl Anstalten treffen} (to
make preparations/to take measures), {\sl treffen} no longer has one of
the meanings translateable with `to hit', `to hurt' or `to meet'; the
noun (plural of {\sl Anstalt}) will in other contexts always have its
literal meaning `institution' but never the meaning of `measure' or
`preparation'.

For collocations that are based on a verb and a noun (usually an
object argument, sometimes this can also be the subject of an
intransitive verb), three types of V--N collocations are distinguished
for German in the literature: 
\begin{itemize}
\item
verbal phrasemes (idioms) (e.g.\
Brundage et al.\ 1992)
\item 
support verb constructions (SVCs) (v.Polenz
1989 or Danlos 1992) 
\item 
collocations in the narrower sense (Hausmann
1989) 
\end{itemize}
The differences between these three types are gradual and it is hard
to find criteria of good selectivity to distinguish collocations from
phrasemes. Although our study is restricted to combinations with
potential support verbs, we will in the following not distinguish
between i) SVCs (e.g.\ {\sl to take into consideration}), ii)
lexicalized combinations with support verbs where the noun has lost
its original meaning and which belong to phrasemes (e.g.\ {\sl to take
a fancy}), and iii) collocational combinations of support verbs with
concrete or non--predicative nouns (e.g.\ {\sl to take a seat}); we
will refer to all these cases as V--N collocations.

Collocations are well suited for statistical corpus studies. According
to Fleischer (1982:63f) the semantics of a collocation in the narrower
sense is ``given by the phrase--external semantics of its components,
but it differs in an unpredictable way from the pure sum of these
component meanings --- however small this difference may be.  [...] 
Language use turns such word combinations into phrase--like
stereotypes. A decisive cause for this [conversion] is the
frequency of occurrence and the probability with which the occurrence
of one component determines the occurrence of the other''\footnote{
``[...] sind Wortverbindungen, deren Gesamtsemantik durch die
wendungsexterne Semantik ihrer Komponenten gegeben ist, die sich aber
doch noch auf nicht voraussagbare Weise --- und sei dies noch so
geringf\"ugig --- von der einfachen Summe dieser Komponentenbedeutungen
unterscheiden. [...] Der Sprachgebrauch wandelt derartige
Wortverbindungen in eine Art phrasenhafter Stereotype um. Ma{\ss}gebend
daf\"ur sind die H\"aufigkeit des Vorkommens und die
Wahrscheinlichkeit, mit der das Auftreten einer Komponente das
Auftreten der anderen determiniert''.
}. 
The high cooccurrence frequency of its components compared to the
relative frequency of the single words thus --- at least partly ---
causes the stereotype status of a collocation, which in turn leads to
its special, phrase--internal semantics. For SVCs and phrasemes this is
even more true because in the course of time they have turned from
stereotypes to completely lexicalised expressions, leading to their
(partly) non--compositional semantics and fixedness.

\section{Related Work}

During the last couple of years, various collocation extraction tools
have been developed for English using frequency thresholds, t--score,
z--score or MI values (see Fontenelle et al.\ (1994) for a survey),
many of them inspired by Berry--Rogghe (1973), Choueka (1988) or the
work of Church and colleagues.  All of them extract some type of
`collocation', but --- with the exception of Smadja (1991b) --- not
much can be found in the literature about the actual quality and
usefulness of the extracted data. Nor have to our knowledge
alternative methods been suggested for German.

Choueka (1988) describes how to automatically extract adjacent word
combinations from English corpora as a preselection of collocation
candidates to ease a lexicographer's search for collocations. He only
uses quantitative selection criteria, no statistical ones, his main
extraction criterion being n--gram frequency with a lower threshold of
at least one occurrence of the collocation in one million words. He
mentions plans to define a `binding degree' on how strong the words of
a collocation attract each other, which would be similar in spirit to
the calculation of MI.

The work on Xtract, described in Smadja and McKeown (1990) and Smadja
(1991a, 1991b, 1993) is along the same lines as that of Church and his
colleagues, trying to improve over a pure frequency approach such as
that of Choueka; but Smadja uses different statistical calculations, a
z--score and variance of distribution, and tagged, lemmatized corpora.
His work goes a step further in also calculating consecutive n--grams
to recognize `rigid noun phrases' and `phrasal templates'. In
addition, syntactic relations determined with shallow parsing can be
taken into account to get a better selection of word combinations.
Using information on syntactic relations, 80\% precision is achieved
compared to 40\% without, recall is only 6\% less than with the
simpler method.

Calzolari and Bindi (1990) use MI to extract compounds, fixed
expressions and collocations from an Italian corpus, but to our
knowledge have not evaluated their results so far.

Grefenstette (1992) has developed Sextant, a system that can
(among other things) produce bigrams for verb and subject, direct or
indirect object combinations, adjective--noun and noun--noun
combinations for English and French. The system itself does not
include additional statistics to determine collocations but the output
can directly be used as input for such calculations.

\section{ Resources and Methods Used in the Study}

Two untagged corpora were used for our study, kindly supplied by the
`Institut f\"{u}r deutsche Sprache' (IdS), Mannheim: the 2.7 million
words `Mannheimer Korpus I' (MK1) which contains approx.\ 73\% fiction
and scientific/philosophical literature and about 27\% newspaper
texts, and the `Bonner Zeitungskorpus' (BZK), a 3.7 million words
newspaper corpus. Except for the test how results could differ for
larger corpora described in section \ref{largecorp}, where the MK1 was
combined with the BZK, the investigation was based on the MK1 on its
own, for technical reasons and also because the pure newspaper corpus
BZK documents a less varied language use (verbs occur more often on
average in the smaller MK1 than in the slightly bigger BZK; cf.\
Breidt 1993).
The statistical calculations were done as described in Church et al.\
(1991), and were performed together with KWIC queries and the creation
of bigrams using tools available at the `Institut f\"{u}r Maschinelle
Sprachverarbeitung', University of Stuttgart\footnote{ We greatfully
acknowledge that the work reported here would not have been possible
without the supplied tools and corpora.}.

\subsection{ Statistical Calculations}

MI is a function well suited for the statistical characterization of
collocations because it compares the joint probability p({\it w1,w2})
that two words occur together within a predefined distance with the
independent probabilities p({\it w1}) and p({\it w2}) that the two
words occur at all in the corpus: 

MI({\it x},{\it y}) = log$_2$( p({\it x},{\it y}) / (p({\it x})p({\it y}) ) 

\noindent
(for a more detailed description see Church et al.\ (1991:120) or
Breidt (1993:18)).
Several methods are possible for the calculation of probabilities
(cf.\ Gale and Church 1990); for our purposes we use the simplest one,
where the frequency of occurrence in the corpus is divided by the size
N of the corpus, p({\it x}) = f({\it x})/N.  Distance is defined as
the window--size in which bigrams are calculated.

MI does not give realistic figures for very low frequencies.  If a
relatively unfrequent word occurs only once in a certain combination,
the resulting very high MI value suggests a strong link between the
words although the cooccurrence might well be simply by chance. To
compensate for this, either a lower frequency bound can be used as an
approximation or better a standard significance test such as a t--test
(e.g.\ Hatch and Farhady 1982).  The t--score indicates whether the
difference between the probability for a collocational occurrence and
the probability for an independent occurrence of the two words is
significant or not.

\subsection{ The `Standard' Method}

In German, common nouns and proper names start with an uppercase
letter (sentence beginnings are changed to lowercase in the corpus)
which makes it possible to extract V--N collocations even from untagged
corpora if the verb is used as the key word. The results give good
indications how promising the retrieval of collocations is with
POS--tagged corpora. We use verbs that can occur in SVCs as key words
because they provide examples for all three types of V--N collocations;
besides, the chosen potential support verbs anyway belong to the most
frequent verbs in the corpus. V--N collocations have been extracted for
the following 16 verbs (no translations are given because they differ
depending on the N argument): \\
{\sl bleiben, bringen, erfahren, finden,
geben, gehen, gelangen, geraten, halten, kommen, nehmen, setzen,
stehen, stellen, treten, ziehen}.

Bigram tables of all words that occur within a certain distance of
these verbs, together with their cooccurrence frequencies, form the
basis for the calculation of MI.  A span of 5 words to the left and
right is said to capture 95\% of significant collocations in English
(Martin et al.\ 1983). So we started with bigram calculations in a
5--word window, but only to the {\em left} of the verb (for a
motivation see next section). We will refer to these with BI5. For
combinations that occur at least 3 times, MI was calculated together
with a t--score.  From these lists, candidates for V--N collocations
were automatically extracted, sorted by MI. For the evaluation, all of
these were manually checked by means of KWIC--listings and classified
by the author w.r.t.\ their collocational status. The classification
was in most cases very obvious. If a combination potentially formed a
collocation but was not used as such in the corpus it did not count; a
couple of times, where some of the usages were indeed collocations and
others not, the decision was made in favour of the predominant case.

No prepositions are extracted in addition to verb and noun, only
bigrams, though MI calculation would also be possible for n--grams,
because bigrams give enough information to test this statistical
approach. In the examples below, prepositions are added manually. 

\section{ Application for German: Some Problems}
\label{german}

Some properties of the German language make the task of extracting V--N
collocations from German text corpora more difficult than for English.
A minor difference concerns the strong inflection of German verbs.
Whereas in English a verb lexeme appears in 3 or 4 different forms
plus one for the present participle, German verbs have 7 to 10 verb
forms (without subjunctive forms) for one lexeme and additional 4 for
the present participle. This has to be considered for the evaluation
of queries based on single inflection forms, because in English more
usages are covered with one verb form than in German.  In section
\ref{lemma} we will see whether it is important for German data to
calculate bigrams based on lemmas.

Another point concerns the variable word order in German (e.g.\
Uszkoreit 1987) which makes it more difficult to locate the parts of a
V--N collocation, regardless whether the corpus is POS--tagged or not.
In a main clause (verb--second order), a noun {\it preceding} a finite
verb usually is the subject, but it may also be a topicalized
complement; in sentences where the main verb occurs at the end
(nonfinite verb or subordinate clause) the preceding noun is mostly a
direct object or other complement, or possibly an adjunct. In a main
clause, a noun {\it to the right} of a finite verb can be any of
subject, object or other argument due to topicalization or scrambling.
The object is located almost at the end of the phrase whereas the main
verb is at verb--second position in the front of the phrase.
Therefore, the assumption that a ``semantic agent [...] is principally
used before the verb'' and a ``semantic object [...] is used after
it'' as suggested by Smadja (1991a:180) does not hold for German.

This has consequences for the choice of a window for bigram
calculations.  Most nouns within 5 words {\it to the right} of a verb
are not its object or prepositional object and therefore are not
potential collocation partners. On the other hand, in subordinate
clauses and in main clauses with a complex tense or modal verb, nouns
{\it to the left} of a verb are quite likely to be its (prepositional)
object.

As a conclusion, with an unparsed corpus we restrict our search to
{\mbox V--N} combinations where the noun precedes the verb within two
to five words, because this is most likely to capture complements of
main verbs in verb--final position.  Furthermore, except for the
experiment simulating lemmatization, we only extract collocations for
verbs in the infinitive form. The infinitive form is used as nonfinite
verb in complex tenses (modal, conditional, future) and is identical
in form with 1st/3rd pers.\ pl.\ present tense and thus covers more
occurrences of the verb than other inflection forms. Besides, the
infinitive is only used in verb--final position.

\section{ Evaluation of the Results}
\label{results}

Below, the top bigrams with {\sl kommen} (come) are shown, and some of
the nonsignificant ones (\mbox{t $\le$ 1.65}), to illustrate MI and
t--scores.

{\small \begin{center}
\begin{tabular}{ll|crrcc}
N + {\sl kommen} & Translation 	   &   f(x,y) &	f(y) &	MI & t--score  &	Coll.\ \\[1ex]
\hline  
\rule[0mm]{0mm}{5mm}(zur) Geltung k.\ & {\small show to advantage}  & 27 &	  96 &	9.86 &	5.19 &	 + \\
(in) Betracht k.\ & {\small to be considered} &	 9 &	  42 &	9.47 &	2.99 &	 + \\
(in) Ber\"{u}hrung k.\ & {\small come into contact} & 4 &  41 &	8.33 &	1.99 &	 + \\
(zur) Anwendung k.\ & {\small to be used} & 4 & 126 &	6.71 &	1.97 &	 + \\
(zu) Tr\"{a}nen k.\ & {\small come to tears} &	 3 &	 107 &	6.53 &	1.70 &	 + \\
(zur) Ruhe k.\ & {\small get some peace} &	 4 &	 216 &	5.93 &	1.95 &	 + \\
(auf d.) Gedanken k.\ & {\small get the idea} & 7 & 403 & 5.84 &	2.58 &	 + \\
(in den) Himmel k.\ & {\small go to heaven} & 3 & 270 & 	5.20 &	1.66 &	 + \\
(zu) Hilfe k.\ & {\small come to aid} &	 4  & 477 &	4.79 &	1.89 &	 + \\
... & & & & & & \\[0.5ex]
(zu) Wort k.\ & {\small get a word in} &	3 & 	647 &	3.94 &	1.57 &	 + \\
Vernunft & {\small reason} &	3 &	 736 &	3.75 &	1.55 &	 -- \\
(in) Frage k.\ & {\small to be possible} &	4 &	1054 &	3.65 &	1.77 &	 + \\
(zur) Welt k.\ & {\small to be born} &	4 &	1900 &	2.80 &	1.60 &	 + \\
Sie & {\small You} &		3 &	2414 &	2.04  &	1.17 &	 -- 
\end{tabular}
\end{center} }

\subsection{ Precision and Recall }

The question how much is extractable fully automatically can be
answered by an evaluation of precision and recall of the described
method as it is done in information retrieval. Following Smadja
(1991a), we define {\em precision} as the number of correctly found
collocations divided by the number of V--N combinations found at all.
{\em Recall} reflects the ratio of the number of correctly found
collocations and the maximal number of collocations that could
possibly have been found. The latter is difficult to determine,
because the total number of collocations occurring in the whole corpus
needs to be known. Thus, we decided to use instead the number of
collocations as determined by the standard method ({\tt BI5}) as the
basis for recall comparisons, i.e.\ 100\% `recall' is set to this
number. Note that this leads to `recall' percentages above 100\%
in case more collocations are extracted than with BI5.

Another possibility had to be discarded, viz to take all collocations
that are mentioned in a collocation dictionary as the maximal number
of valid collocations: a comparison with Agricola (1970) or Drosdowski
(1970) is not really possible because the collocations found in the
corpus are not a subset of those mentioned in the dictionaries. Only
22 of the 43 collocations found with the lemma {\sl bring--} in the MK1
({\tt BI5}) belong to the 135 combinations mentioned in the lexical
entry for {\sl bringen} in Agricola (1970).  Of the remaining 21 in
the MK1, 9 can be found in the corresponding noun entries, and 12 do
not appear at all though they are `significant' collocations, e.g.\
{\sl Klarheit bringen} (to clarify), {\sl zur Entfaltung bringen} (to
develop), {\sl zur Wirkung bringen} (to bring the effect), {\sl in
Schwierigkeiten bringen} (to create difficulties), {\sl ins
Gespr\"{a}ch bringen} (to bring into discussion).

\subsection{ Results of the Standard Method}

Frequency of the infinitive form of the 16 verbs ranges from 832 ({\sl
kommen}) to 117 ({\sl gelangen}). The number of V--N combinations
occurring at least 3 times varies from 46 ({\sl bringen}) to 6 ({\sl
erfahren, gelangen, geraten, treten}), precision from 100\% ({\sl
geraten, ziehen}) to 33\% ({\sl erfahren}).  Average figures are
presented in table \ref{averages} below, labeled {\tt BI5 Inf}.  If
non--significant combinations are omitted with a t--test ({\tt BI5/t
Inf}), the average number of extracted collocations is minimally
lower, and precision rises slightly. With a threshold of MI $\ge$ 6,
precision goes up to 81.6\% with a loss in recall of 10\%.

For {\sl bringen}, the approximate absolute number of collocations in
the MK1 was manually determined. Out of 585 different V--N
combinations, 71 can generously be classified as collocations. Of
these, 31 are automatically extracted with BI5, so absolute recall in
this case is 43.7\% (precision 67.4\%).

\subsection{ Experiment 1: Variation of Window--Size}

Except in cases where a post--modifier follows the object or where the
object is topicalized, a direct object or prepositional object is more
likely to be directly to the left of the verb rather than within a
couple of words.  So we reduced window--size to 2 words to the left of
the verb which allows one word in between, e.g.\ infinitival {\sl zu}
(to). As shown in table \ref{averages} for {\tt BI2 Inf}, precision
rises about 15\%, but with a recall of 72.1\%, because those
collocations where other arguments or post modifiers occur between N
and V are no longer captured.  This leads to the conclusion that for
German, unless syntactic relations can be determined, a smaller window
is preferable to improve a correct detection of preceding object
arguments and to exclude unrelated nouns. Taking again only
significant combinations and using an MI threshold of 6 ({\tt BI2/t
Inf MI}) precision rises to 90\%, with a low recall of 57\%.

\vspace{-2mm}
\begin{table}[h] 
\caption{\label{averages} Average Figures of Variant Methods for the 16 Verbs}
{\small \begin{center}
\vspace{-2mm}
\begin{tabular}{lcl}
\hline
Bigrams \& Filter	&	{\O} Precision \% &	{\O} Recall \% \\	
\hline
BI5 Inf 			&  66.3 & 100 (def.) \\
BI5/t Inf 			&  71.6 & 95.8 \\
BI5/t Inf, MI			&  81.6 & 84.7 \\[0.8ex]
BI2 Inf 			&  81.2 & 72.1 \\
BI2 Inf, no subj 		&  85	& 72.1 \\
BI2/t Inf 			&  83.1 & 70.0 \\
BI2/t Inf, MI			&  90.1	& 57 \\[0.8ex]
BI2 Lemma 			&  59.8	& 114.7 \\
BI2 Inf+Part			&  78.9	& 99.96 \\[0.8ex]
BI2 Inf Mk+Bz			&  72.3 & 186 \\
BI2 Inf Mk+Bz, MI 		&  87.6 & 154.2 \\
BI2 Inf Mk+Bz, f$\ge$5 		&  89.4 & 95.9 \\
BI2 Inf Mk+Bz, MI, f$\ge$5	&  97.8 & 87.7 \\
\hline
\end{tabular}
\end{center} }
\vspace{-5mm}
\end{table}
\vspace{-5mm}

\subsection{ Experiment 2: Simulating Lemmatizing}
\label{lemma}

Because no lemmatizing program was available we used an additional
program on top of the bigram calculations for the inflected forms. In
order to keep the amount of V--N combinations within a magnitude that
could still be checked manually for correctness, we restricted search
to a 2--word window to the left. V--N combinations that occurred less
than two times with a single inflection form of the verb were sorted
out. All inflection forms were added up except 1st pers.\ sg.\ and 2nd
pers.\ sg./pl.\ because these were so rare that they could be ignored.
The average results are again presented in table \ref{averages} ({\tt
BI2 Lemma}); the number of extracted collocations is maximal, but
precision is the lowest of all. Precision ranges from 88.2\% ({\sl
setzen}) to 33.3\% ({\sl gehen}), recall from 166.7\%\footnote{ Recall
figures are above 100\% because the absolute number of collocations
found is higher than for {\tt BI5 Inf}, the basis for our recall
calculations.} ({\sl setzen}) to 50\% ({\sl erfahren}).  If
collocations for the infinitive and past participle only are extracted
({\tt BI2 Inf+Part}), recall is as good as with {\tt BI5 Inf}, with an
improved precision of 78.9\%.

Regarding lemmatization our study shows that one gets more
collocations, but at the expense of more uninteresting combinations as
well. One explanation for this is that 3rd pers.\ sg.\ present/past
and 1st/3rd pers.\ pl.\ past verbforms occur to the right of their
noun argument only in subordinate clauses; the nonfinite form,
identical with 1st/3rd pers.\ pl.\ present, and the past participle
additionally occur in verb--final position to the right of the noun
argument in main clauses with a finite auxiliary or modal verb and in
infinitive clauses. Therefore, with an unparsed corpus, only the
infinitive and past participle should be used for extractions.

\subsection{ Experiment 3: Varying Corpus Size}
\label{largecorp}

V--N combinations within 2 words to the left of the infinitive were
also calculated for a larger corpus consisting of the MK1 and BZK
together (6.4 mio word tokens).  The number of V--N collocations found
with {\tt BI2 Inf} is almost twice as big (186\% `recall'), with a
slightly lower precision than for the small corpus. So, larger corpora
considerably improve recall.  Raising the frequency threshold to five
improves precision but drastically cuts down recall to its half. With
an MI threshold instead of frequency results are much better, almost
the same precision with 154\% recall, so using an MI measure rather
than pure frequencies has clear advantages. If both are combined, less
than 3 out of 100 combinations are wrongly extracted, of course with a
low recall of less than half of the unfiltered {\tt BI2 Inf}.
Considering that the corpus is untagged, this is quite a good result
for the automatic acquisition of collocations.

\subsection{Experiment 4: Simulating Syntactic Tagging}
\label{subj}

In main clauses with simple tense, most often the subject is to the
left of the inflected verb, so it is likely that for verbs in 1st/3rd
pers.\ pl.\ present (identical with the infinitive form), the nouns
extracted to the left are in fact subjects of the verb and thus do not
form a collocation with it. Note that a POS--tagged but unparsed corpus
would not solve this problem because the correct distinction of
infinitives and inflected 1st/3rd pers.\ pl.\ present tense forms is
one of the difficulties in automatic POS--tagging.  In order to see how
much the precision could possibly be improved by determining syntactic
relations as done by Smadja (1991a,b) for English, we conducted
another test where we manually excluded from {\tt BI2 Inf} those
wrongly extracted combinations in which the nouns were in fact used in
subject position of the verb.

Precision would rise to 85\% on average if one could consider
syntactic relations for the extraction of V--N collocations ({\tt `BI2
Inf, no subj'} in table \ref{averages}).  In that case, extraction
would of course no longer need to be limited to a 2--word window to the
left, so recall results should be better, too. Interestingly,
improvements are not as big as expected, so quite a few of the nouns
must be part of a post--modifier or adjunct. Results with parsed data
should thus be better.

These results point in the same direction as Smadja's who reports an
improvement from 40 to 80\% precision if syntactic relations are
considered, with a 94\% recall of all collocations that had been found
regardless of syntactic relations.  However, this cannot as easily be
achieved in a large scale for German due to the complicated parsing
techniques necessary for the varying word order.

\section{ Conclusions and Outlook}

The task of extracting V--N collocations can be split into two
sub--tasks: 
\begin{itemize}
\item[(i)] the extraction of possible candidates
\item[(ii)] the filtering of undesired combinations. 
\end{itemize}
For (i), a general basic requirement is a POS--tagged corpus, unless
verbs or adjectives are given as key words and combinations with nouns
are to be extracted. With an unparsed corpus, restricting the
extraction to nouns within 2 words to the left of an infinitive or
past participle yields the best results for task (i) with a precision
of around 80\% ({\tt BI2 Inf+Part}). Lemmatization is not very useful
unless parsed corpora are available ({\tt BI2 Lemma}).  Larger corpora
improve recall without a serious decline in precision, though some
additional noise seems to come along with the bigger amount of data
({\tt BI2 Inf Mk+Bz}).  Of course, the determination of syntactic
relations would open up still better possibilities to find collocation
candidates.

MI and t--score thresholds work quite satisfactorily as filters for
task (ii), but at least half of the actually occurring collocations
are filtered out, too. MI sorts the extracted combinations in such a
way that the collocations are the better the higher the MI--score is
(with a few exceptions which often reflect highly significant, but
linguistically uninteresting word combinations from one of the texts;
this could hopefully be avoided with a more balanced corpus).
Possibly, the likelyhood ratio method suggested by Dunning (1993)
might be an interesting alternative to MI.

Very high precision rates that do not cut down too much on recall, an
indispensible requirement for lexical acquisition, can best be
achieved for German with parsed corpora. As long as such resources are
not widely available, using a large corpus and high MI and t--score
filters should produce satisfactory results ({\O} 97.8\% precision,
{\tt `BI2 Inf Mk+Bz,MI,f$\ge$5'}).  A good lexicographical support
where a better recall is important can be provided with less
restrictive filters. The usefulness of the resulting lists should not
be underestimated both for manually built NLP lexica and for printed
dictionaries.  The simple variant {\tt BI2 Inf} together with MI is
successfully used in the EEC--funded COMPASS project (LRE 62--080) to
support lexicographic work on augmenting a large German--English
bilingual dictionary with missing collocations and idioms and checking
the relevance of already included multi--word expressions.

Furthermore, the described approach seems to be a good method for
corpora with texts from restricted domains, where a special
terminology is used which will thus show up strongly against `normal'
combinations.

Work is currently in progress to calculate trigrams and larger n--grams
to check for prepositions in SVCs or for specific (or no) determiners
for phrasemes. This will give indications to distinguish SVCs and
lexicalized, phraseological SVCs from other collocations. In addition,
we plan to consider the variation in span position of the noun within
the searched window in order to distinguish fixed phrasemes from
flexible ones. 

\section*{References}
{\small 
\parskip4pt plus 2pt minus 1pt
\noindent
Agricola, E., H.\ G\"{o}rner, R.\ K\"{u}fner (eds.) (1962/1970).  {\sl
W\"{o}rter und Wendungen. W\"{o}rterbuch zum deutschen
Sprachgebrauch.} Leipzig, M\"unchen: Verlag Enzyklop\"{a}die.

\noindent
Berry--Rogghe, G.\ (1973). The computation of collocations and their
relevance in lexical studies. In: Aitken, A.\ J., R.\ W.\ Bailey, N.\
Hamilton--Smith (eds.). {\sl The Computer and Literary Studies}. Edinburgh:
University Press. 103--112.

\noindent
Breidt, E.\ (1993). {\sl Extraktion von Verb--Nomen--Verbindungen aus dem
Mannheimer Korpus I}. SfS--Report 03--93. University of T\"{u}bingen.

\noindent
Brundage, J., M.\ Kresse, U.\ Schwall, A.\ Storrer (1992). {\sl Multiword
lexemes: a monolingual and contrastive typology for NLP and MT}.
IWBS--Report 232, September 1992. IBM Germany, Scientific Centre Heidelberg.

\noindent
Calzolari, N., R.\ Bindi (1990). Acquisition of lexical information
from a large textual italian corpus. {\sl 13th COLING 1990}, Helsinki.
54--59.

\noindent
Choueka, Y.\ (1988). Looking for needles in a haystack, or: locating
interesting collocational expressions in large textual databases.
{\sl Proceedings of the RIAO}. 609--623.

\noindent
Church, K.\ W., P.\ Hanks (1989). Word Association Norms, Mutual
Information and Lexicography. {\sl 27th ACL}, Vancouver. 76--83. 

\noindent
Church, K.\ W., W.\ A.\ Gale, P.\ Hanks, D.\ M.\ Hindle (1991). Using
statistics in lexical analysis. In: Zernik, U.\ (ed.). {\sl Lexical
acquisition: exploring on--line resources to build a lexicon}.
Hillsdale, NJ.

\noindent
Danlos, L.\ (1992). Support verb constructions. Linguistic properties,
representation, translation. {\sl Journal of French Linguistic Study, Vol.\
2, No.\ 1}. CUP.

\noindent
Drosdowski, G.\ et al.\ (eds.) (1970). {\sl Duden Stilw\"{o}rterbuch der
deutschen Sprache: Die Verwendung der W\"{o}rter im Satz}. 6th
completely revised and extended edition. Mannheim.

\noindent
Dunning, T.\ (1993). Accurate Methods for the Statistics of Surprise
and Coincidence. {\sl Computational Linguistics, Vol.\ 19, No.\ 1}. 61--74.

\noindent
Fleischer, W.\ (1982). {\sl Phraseologie der deutschen Gegenwartssprache}.
Leipzig.

\noindent
Fontenelle, Th.\, W.\ Bruls, L.\ Thomas, T.\ Vanallemeersch, J.\
Jansen (1994). {\sl Survey of collocation extraction tools}.
Deliverable D--1a, MLAP--Project 93--19 DECIDE. University of Li\`ege,
Belgium, July 1994.

\noindent
Gale, W., K.\ W.\ Church (1990). What's wrong with adding one?
{\sl IEEE Transactions on Acoustics, Speech and Signal Processing}.

\noindent
Grefenstette, G.\ (1992). Use of Syntactic Context to Produce Term
Association Lists for Text Retrieval. 
{\sl 15th ACM SIGIR Conference on
Research and Development in Information Retrieval}. Copenhagen,
Denmark. 89--97.


\noindent
Hatch, E., H.\ Farhady (1982). {\sl Research design and statistics for
applied linguistics}. Rowley.

\noindent
Hausmann, F.\ J.\ (1989). Le dictionnaire de collocations. In:
Hausmann, F.\ J.\ et al.\ (eds.). {\sl Dictionaries: an international
handbook for lexicography. Part I}. HSK 5.1. Berlin: De Gruyter. 1010--1019.

\noindent
Martin, W., B.\ Al, P.\ van Sterkenburg (1983). On the processing of a
text corpus. In: Hartmann, R.\ R.\ K.\ (ed.). {\sl Lexicography: principles
and practice}. London. 77--87.

\noindent
v.Polenz, P.\ (1989). Funktionsverbgef\"{u}ge im allgemeinen
einsprachigen W\"{o}rterbuch. In: Hausmann, F.\ J.\ et al.\ (eds.).
{\sl Dictionaries: an international handbook for lexicography. Part I}. HSK
5.1. 882--887.

\noindent
Sinclair, J.\ M.\ (1966). Beginning the study of lexis. In: Bazell,
C.\ E.\ et al.\ (eds.) (1966). {\sl In memory of J.\ R.\ Firth}. London.
410--430.

\noindent
Smadja, F.\ A., K.\ R.\ McKeown (1990). Automatically extracting and
representing collocations for language generation. {\sl 28th ACL
1990}. 252--259.

\noindent
Smadja, F.\ A.\ (1991a).  Macrocoding the lexicon with co--occurrence
knowledge. In: Zernik, U.\ (ed.). {\sl Lexical acquisition: exploring
on--line resources to build a lexicon}. Hillsdale, NJ.

\noindent
Smadja, F.\ A.\ (1991b). From n--grams to collocations: an evaluation
of Xtract. {\sl 29th ACL, Berkeley, CA}. 279--284.

\noindent
Smadja, F.\ A.\ (1993). Retrieving collocations from text: XTRACT.
{\sl Computational Linguistics}, Vol.\ 19, No.\ 1. 143--177.

\noindent
Uszkoreit, H.\ (1987). {\sl Word order and constituent structure}. CSLI
Lecture Notes 8.  

} 
\end{document}